\documentclass[letterpaper, 10 pt, conference]{ieeeconf}  % Comment this line out if you need a4paper

\IEEEoverridecommandlockouts                              % This command is only needed if 
\usepackage{times}
\usepackage{multicol}
\usepackage{graphicx}
\usepackage{subfigure}
\usepackage{caption}
\usepackage{csquotes}
\usepackage{amsmath}
\usepackage{multirow}
\usepackage{booktabs}
\usepackage[bookmarks=true]{hyperref}
\usepackage{algorithm}
\usepackage{algorithmic}
\usepackage{url}            % simple URL typesetting
\usepackage[table,xcdraw]{xcolor} 
\definecolor{Blue}{HTML}{e5f4fc}
\definecolor{Pink}{HTML}{f4e3f4}
\definecolor{LightBlue}{RGB}{81, 131, 251}

\title{
    \LARGE \bf
   \raisebox{-0.3ex}{\includegraphics[height=1.2em]{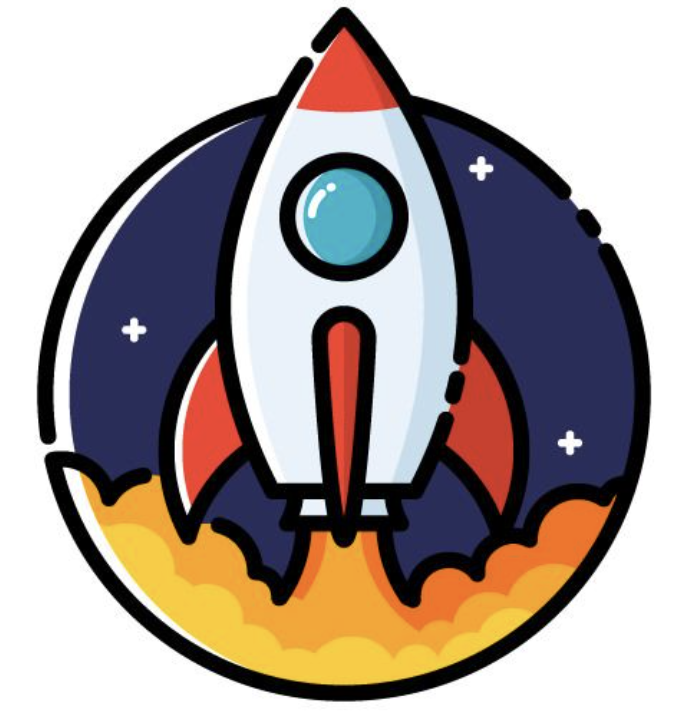}}\textbf{SimLauncher}: \underline{Launch}ing Sample-Efficient Real-world Robotic Reinforcement Learning via \underline{Sim}ulation Pre-training
  
}

\author{Mingdong Wu$^*$, Lehong Wu$^*$, Yizhuo Wu$^*$, Weiyao Huang, Hongwei Fan, Zheyuan Hu, \\ Haoran Geng,
Jinzhou Li, Jiahe Ying, Long Yang, Yuanpei Chen, Hao Dong% <-this % stops a space
\thanks{Mingdong Wu, Lehong Wu, Yizhuo Wu, Weiyao Huang, Hongwei Fan, Jinzhou Li, Jiahe Ying, and Long Yang are with the Center on Frontiers of Computing
Studies, School of Computer Science, Peking University, also with PKUAgibot Lab. 
Haoran Geng is with the University of California, Berkeley.
Zheyuan Hu is with the Robotics Institute at Carnegie Mellon University.
Yuanpei Chen is with the PKU-Psibot Joint Lab.
}
\thanks{*indicates equal contribution.}% <-this % stops a space
\thanks{Corresponding to hao.dong@pku.edu.cn.}% <-this % stops a space
}

\begin{document}

\def\eg{\emph{e.g}.} \def\Eg{\emph{E.g}.}
\def\ie{\emph{i.e}.} \def\Ie{\emph{I.e}.}
\def\cf{\emph{c.f}.} \def\Cf{\emph{C.f}.}
\def\etc{\emph{etc}.} \def\vs{\emph{vs}.}
\def\wrt{w.r.t. } \def\dof{d.o.f. }
\def\etal{\emph{et al}. }

\maketitle

\vspace{-5pt}

%%%%%%%%%%% figure 1 teaser %%%%%%%%%%%%
\begin{figure*}[ht]
\begin{center}
% \vspace{-15pt}
\includegraphics[width=0.98\linewidth, trim=0cm 0cm 0cm 0cm, clip]{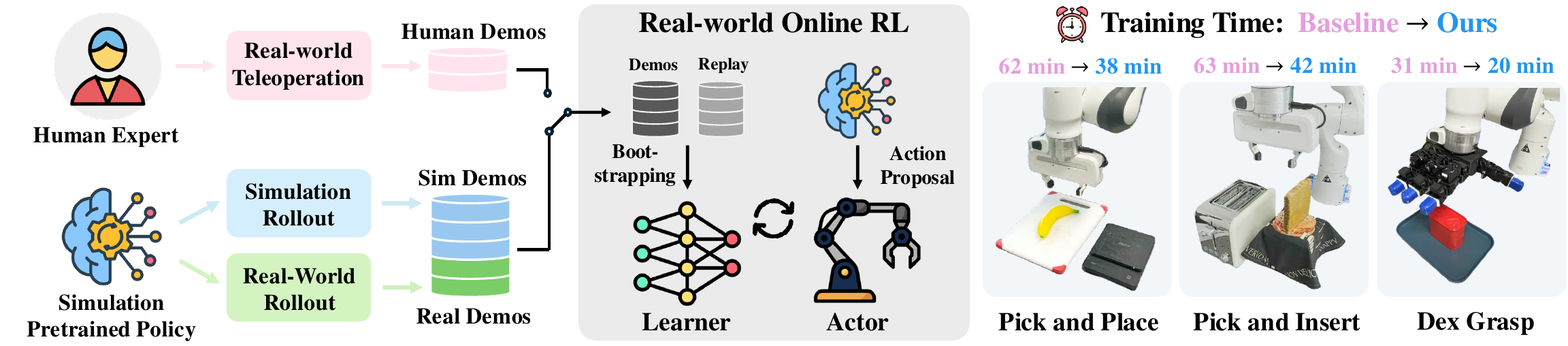}
\end{center}
\vspace{-10pt}
\caption{\textbf{Illustration of our motivation.} Given a simulation-pretrained policy, SimLauncher leverages simulated and real-world rollouts as demonstrations for critic bootstrapping and incorporates the policy for action proposal. SimLauncher significantly improves the sample efficiency of real-world RL compared with conventional RL methods using human-collected data.
\vspace{-20pt}
}

\end{figure*}
%%%%%%%%%%%%%%%%%%%%%%%%%%%%%%%%%%%%

%%%%%%%%%%%%%%%%%%%%%%%%%%%%%%%%%%%%%%%%%%%%%%%%%%%%%%%%%%%%%%%%%%%%%%%%%%%%%%%%
\begin{abstract}

Autonomous learning of dexterous, long-horizon robotic skills has been a longstanding pursuit of embodied AI. 
Recent advances in robotic reinforcement learning (RL) have demonstrated remarkable performance and robustness in real-world visuomotor control tasks. However, applying RL in the real world faces challenges such as low sample efficiency, slow exploration, and significant reliance on human intervention.
In contrast, simulators offer a safe and efficient environment for extensive exploration and data collection, while the visual sim-to-real gap, often a limiting factor, can be mitigated using real-to-sim techniques.
Building on these, we propose SimLauncher, a novel framework that combines the strengths of real-world RL and real-to-sim-to-real approaches to overcome these challenges.
Specifically, we first pre-train a visuomotor policy in the digital twin simulation environment, which then benefits real-world RL in two ways: (1) bootstrapping target values using extensive simulated demonstrations and real-world demonstrations derived from pre-trained policy rollouts, and 
(2) Incorporating action proposals from the pre-trained policy for better exploration.
We conduct comprehensive experiments across multi-stage, contact-rich, and dexterous hand manipulation tasks. Compared to prior real-world RL approaches, SimLauncher significantly improves sample efficiency and achieves near-perfect success rates.
We hope this work serves as a proof of concept and inspires further research on leveraging large-scale simulation pre-training to benefit real-world robotic RL. 
% More demonstrations can be viewed at \url{https://simlauncher-web.vercel.app/}.

\end{abstract}

\def\isMainDocument{}

\section{Introduction}
% Flag: finish intro by 2.27
% 1. Big Picture
For decades, researchers have been captivated by the pursuit of embodied AI that could seamlessly generalize across diverse tasks, demonstrate dexterous manipulation, and achieve flawless performance upon deployment. 
While significant strides have been made in generalizable learning~\cite{xie2024decomposing, pmlr-v164-jang22a, ju2024robo, pumacay2024colosseum} and dexterous manipulation~\cite{rajeswaran2017learning, chen2022towards, zhao2024aloha}, the challenge of robust manipulation—ensuring consistent success across real-world conditions—remains an underexplored frontier, yet one that is indispensable for the industrialization of embodied intelligence. 

% 2. Entry Point: Real World RL and its limitations
While directly applying reinforcement learning to robotic tasks in real-world settings is a conceptually promising approach for acquiring robust robot policies, training policies in physical environments remains both unsafe and costly. 
Even with carefully designed, compliant controllers to address safety concerns~\cite{luo2019reinforcement}, the challenges of exploration still hinder sample efficiency. 
Recent advancements~\cite{luo2024precise, luo2024serl} have demonstrated that integrating human-collected demonstrations to bootstrap the critic or using human interventions to guide exploration can significantly enhance learning efficiency, enabling near-optimal visuomotor policies for a diverse range of precise and dexterous skills. However, these methods rely on labor-intensive data collection and manual interventions, making them costly and difficult to scale.

% 3. motivated by sim generalist and decreasing sim-to-real gap
How can we enhance data coverage for real-world robotic reinforcement learning bootstrapping and improve exploration while minimizing human labor costs?
One potential approach is to leverage simulator. 
On the one hand, significant advancements have been made in automatically generating large-scale robotic task trajectories in simulation~\cite{duan2024manipulate,wang2023robogen,wang2023gensim,hua2024gensim2,mandlekar2023mimicgen,jiang2024dexmimicgen,garrettskillgen,garrett2024skillmimicgen,nasiriany2024robocasa}, making it increasingly feasible to develop a simulation-pretrained generalist policy. 
On the other hand, Real-to-Sim-to-Real techniques have been extensively studied in recent works~\cite{torne2024reconciling,qureshi2024splatsim, lou2024robo, han2025re,wu2024rl,lou2024robo, patel2025real} and have shown promising results in mitigating the Sim-to-Real gap~\cite{huang2023went,he2023bridging} and building a digital twin more efficiently~\cite{makoviychuk2021isaac, Zhou_2024, zakka2025mujocoplayground}.
These insights lead us to the central research question of this study:  
\begin{displayquote}  
    \textit{How can simulation-pretrained policies, together with a digital twin, improve the sample efficiency of real-world robotic reinforcement learning?}  
\end{displayquote}   
To this end, we conduct a proof-of-concept study that restricts our task scenarios to fixed objects and backgrounds and pre-trains a specialized policy within the digital twin.

% 4. Our method
We propose \textbf{SimLauncher}, a simple yet effective vision-based reinforcement learning framework that leverages a simulated pre-trained policy along with a digital twin to improve real-world RL’s sample efficiency. 
We simulated rollouts generated by the pre-trained policy as demonstrations for critic bootstrapping—providing hundreds of trajectories that significantly expand state coverage. 
To prevent the critic from over-exploiting the task-irrelevant features arising from differences between simulated demonstrations and the real-world replay buffer, we further incorporate a limited number of real-world rollouts from the pre-trained policy to regularize training. 
Moreover, following IBRL~\cite{hu2023imitation}, we use the pre-trained policy to accelerate exploration by selecting the action with the higher critic score between those proposed by the pre-trained and the RL policy. 
Unlike existing methods that integrate state-based simulation digital twins with real-world RL, SimLauncher adopts a vision-based setting, offering enhanced robustness and adaptability.

% 5. Experiments and claims, tbd after finished the exp section
We conduct extensive experiments on three challenging real-world tasks: a multi-stage task, a precise manipulation task, and a dexterous hand manipulation task with high-dimensional action space. SimLauncher outperforms conventional hybrid RL baselines that rely on human-collected demonstrations.
Our analysis suggests that simulated data alone can effectively support bootstrapping.
% Our analysis suggests that simulated data effectively supports bootstrapping despite the sim-to-real gap.

% Ablation studies reveal that all ablated versions perform significantly worse than the full method, underscoring the importance of each design choice. 

% Simulation Data Generation: 
% VoxPoser~\cite{huang2023voxposer}
% Rekep~\cite{huang2024rekep}
% Manipulate Anything~\cite{duan2024manipulate}
% OmniManip~\cite{pan2025omnimanip}
% Eureka~\cite{ma2023eureka}
% RoboGen~\cite{wang2023robogen}

% GenSim~\cite{wang2023gensim}
% GenSim2~\cite{hua2024gensim2}
% MimicGen~\cite{mandlekar2023mimicgen}
% DexMimicGen~\cite{jiang2024dexmimicgen}
% SkillGen~\cite{garrettskillgen}
% SkillMimicGen~\cite{garrett2024skillmimicgen}
% RoboCasa~\cite{nasiriany2024robocasa}
% DiffGen(cewu)~\cite{taneja2008diffgen}
% IntervenGen~\cite{hoque2024intervengen}

\vspace{-5pt}
\section{Related Works} 
\label{sec:related_works}
\vspace{-3pt}
\subsection{Real-World Robotic Reinforcement Learning}
Real-world robotic reinforcement learning (RL) requires sample-efficient algorithms for high-dimensional inputs like onboard perception, with easy reward and reset specification. Several methods have shown efficient learning from scratch in the real world~\cite{luo2024serl, luo2024precise, hu2023imitation, ball2023efficient} or fine-tuning generalist policies~\cite{guo2025improving, chen2025conrft}. Recent advances in reset-free learning~\cite{hu2023reboot, gupta2021reset, xu2023dexterous, zhu2020ingredients} aim to minimize human intervention.
While prior work has focused on improving sample efficiency in off-policy RL via offline pretraining~\cite{nair2020awac, yang2024robot, zhou2024efficient}, hybrid RL~\cite{ball2023efficient, hu2023imitation}, or leveraging foundation model priors~\cite{ye2023reinforcement}, our approach accelerates real-world RL by incorporating simulation. 
In this line, previous studies have used simulation-trained value functions for exploration~\cite{yin2025rapidly}, simulation-pretrained actors for exploration efficiency~\cite{wagenmaker2024overcoming}, or fine-tuning pre-trained actors with significant unlearning periods~\cite{zhang2023cherry}.
In contrast, we propose leveraging a simulation-pretrained policy for both exploration and generating demonstration rollouts to bootstrap critic training. To our knowledge, this is the first bootstrapping-based approach that uses simulation to enhance real-world RL sample efficiency. Furthermore, unlike prior state-based methods, we consider a vision-based setting which improves robustness and generalizability across diverse real-world scenarios.

\vspace{-3pt}
\subsection{Real-to-Sim-to-Real Approaches for Policy Learning}
\vspace{-3pt}

Sim2Real transfer is a promising and widely explored approach for robotic policy learning~\cite{zhao2020sim}~\cite{qi2023hand}~\cite{wang2024lessons}. However, the visual and physical discrepancies between simulation and the real world present significant challenges~\cite{zhao2020sim}~\cite{singh2024dextrah}.

To address the visual gap, \cite{torne2024reconciling} propose the first Real2Sim2Real framework, utilizing augmented demonstrations synthesized from a digital twin to enhance the robustness of behavioral cloning. To reduce the burden of reconstructing a digital twin, ACDC~\cite{dai2024automated} introduces a framework for retrieving digital assets that exhibit similar geometric and semantic affordances to the target task scenario.

Inspired by the promising results of 3D Gaussian Splatting~(3DGS), recent studies have incorporated 3DGS for Real2Sim reconstruction~\cite{qureshi2024splatsim, lou2024robo, han2025re}, enabling training of RGB-based visuomotor policies via reinforcement learning~\cite{wu2024rl}, reconstructing articulated robot arms~\cite{lou2024robo}, and augmenting real-world data. Additionally, another line of research explores leveraging Real2Sim for scalable, high-quality data generation~\cite{mu2024robotwin, patel2025real, ye2025video2policy}, facilitating future research on training generalist policies in simulation.

Our work integrates 3DGS-based Real2Sim to train a visuomotor policy and generate simulated demonstrations for the target task. Unlike previous approaches, we utilize the trained policy for action proposal in real-world RL, while the generated demonstrations are used to bootstrap the critic.

%%%%%%%%%%%%%%%%%%%%%%%%%%%%%%%%%%%%%%
\vspace{-5pt}
\section{Method} 
\label{sec:method}
\vspace{-5pt}

% \mingdong{we will simplify or remove this if we do not have enough pages. Method still needs major revision.}

%%%%%%%%%%% figure 2 method %%%%%%%%%%%%
\begin{figure*}[t]
\begin{center}
% \vspace{-15pt}
\includegraphics[width=0.93\linewidth, trim=0cm 0cm 0cm 0cm, clip]{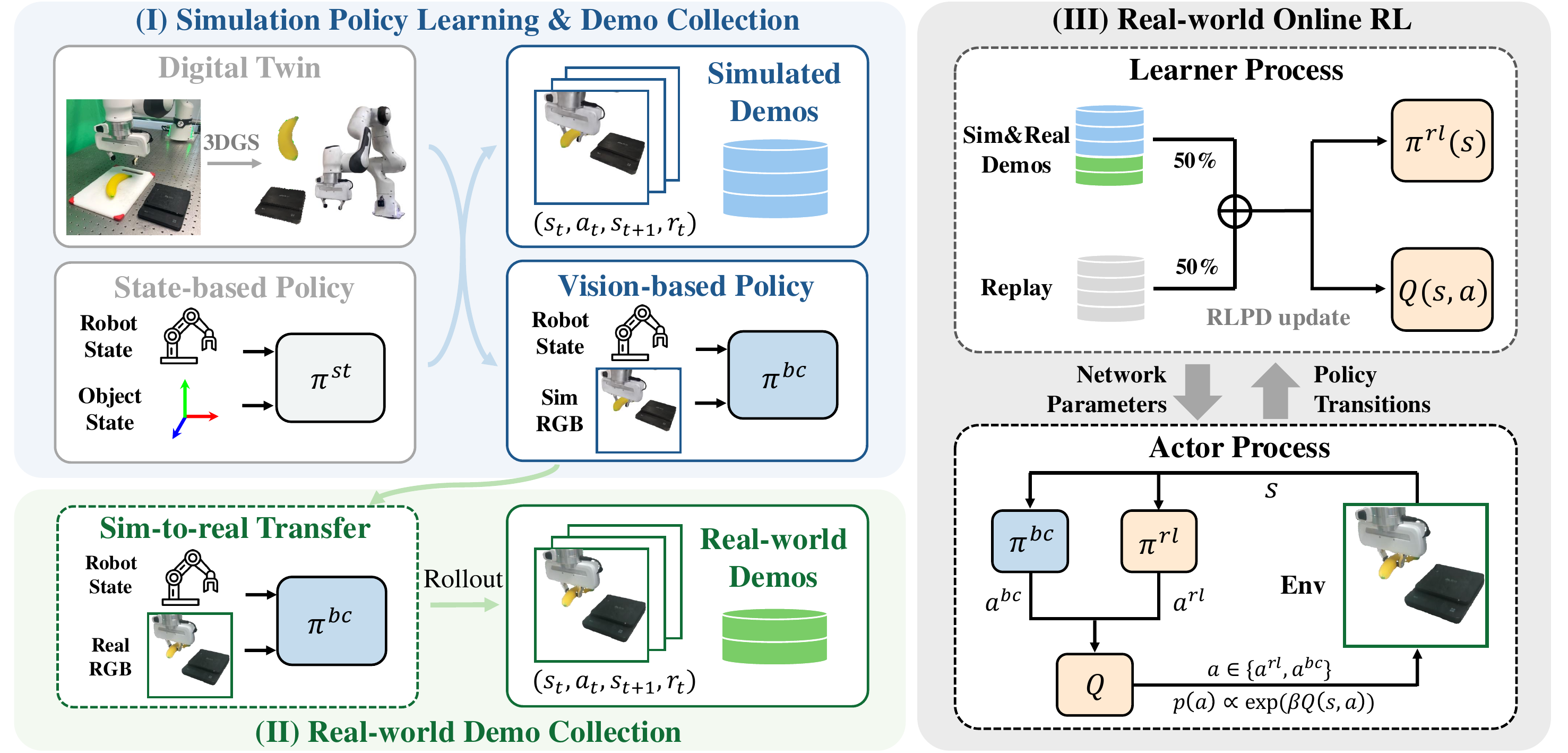}
\end{center}
\vspace{-10pt}
\caption{\textbf{Overview of SimLauncher.} In simulation, we collect simulated demos and train a vision-based policy for each task. We then rollout the pre-trained policy in the real world to collect real demos. 
The simulated and real-world demos are used for critic bootstrapping in real-world RL. The pre-trained policy also provides action and bootstrap proposal for the actor.
}
\vspace{-17pt}
\end{figure*}

\vspace{-3pt}
\subsection{Overview}
\label{subsec: overview}
\textbf{Preliminaries.} We model our robotic tasks as Markov Decision Processes $\mathcal{M} = \{\mathcal{S}, \mathcal{A}, \rho, \mathcal{P}, r, \gamma\}$, where $\mathcal{S}$ is the observation space, $\mathcal{A}$ is the action space, $\rho(s_0)$ is the distribution over initial states, $\mathcal{P}$ is the transition probability function, $r: \mathcal{S} \times \mathcal{A} \rightarrow \{0,1\}$ is the sparse reward function, and $\gamma$ is the discount factor. 
Reinforcement learning (RL) aims to find the optimal policy $\pi$ that maximizes the cumulative reward $E\left[\sum_{t=0}^{T}\gamma^tr(s_t,a_t)\right]$.
The core algorithm of our method is RLPD~\cite{ball2023efficient}, a hybrid RL algorithm that bootstraps on prior data.
% to improve sample efficiency and exploration. 
For each batch, RLPD performs 50/50 sampling from the replay buffer and the demo buffer. The following objectives are optimized for the parametric critic $Q_{\varphi}(s,a)$ and actor $\pi_\theta(a|s)$: 
\vspace{-8pt}
\begin{equation}
\label{eq:q-target}
   y = r + \gamma E_{a' \sim \pi_{\theta}} \left[ Q_{\tilde{\varphi}}(s', a') \right],
\end{equation}
\vspace{-15pt}
\begin{equation}
    \mathcal{L}_Q(\varphi) = E_{s, a, s'} \left[ \left( Q_{\varphi}(s, a) - y\right)^2\right],
\end{equation}
\vspace{-15pt}
\begin{equation}
     \mathcal{L}_{\pi}(\theta) = -E_s \left[ E_{a \sim \pi_{\theta}} \left[ Q_{\varphi}(s, a) - \alpha \,\text{log}\,\pi_\theta(a|s)\right] \right],
\end{equation}
where $Q_{\tilde{\varphi}}(s,a)$ is the target Q network, and $\alpha$ is the entropy temperature controlling policy randomness. 

\begin{figure*}[t]
\begin{center}
% \vspace{-10pt}
\includegraphics[width=0.95\linewidth, trim=0cm 0cm 0cm 0cm, clip]{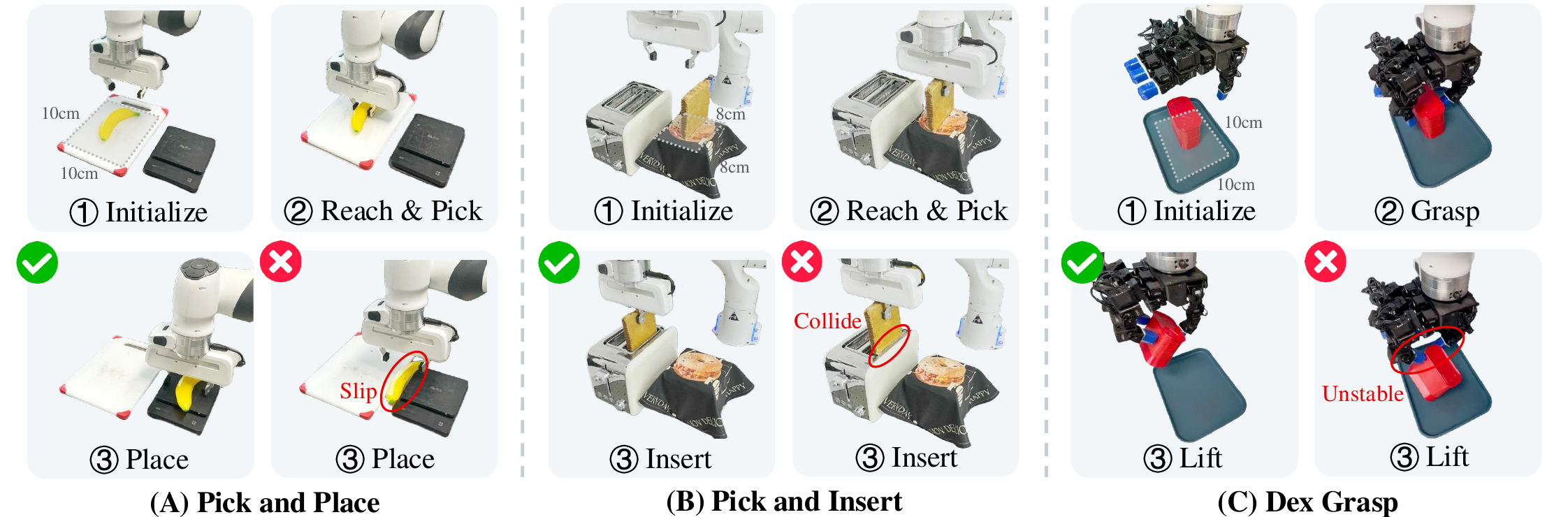}
\end{center}
\vspace{-7pt}
\caption{
\textbf{Task illustrations, initialization ranges, and common failure modes.} (A) Pick and Place. This task involves relocating a banana to an electronic scale. Common failure includes the banana slipping off the scale. (B) Pick and Insert. This task involves grasping a toast and inserting it into
the correct slot of a toaster. 
Common failure includes the toast getting stuck on the toaster edge. 
(C) Dex Grasp. This task involves coordinating multiple fingers to achieve a force closure on a can and lift it by 5 cm. Common failure includes the can slipping off during lifting or being knocked over by the hand. }
\label{fig:task_illustration}
\vspace{-18pt}
\end{figure*}

\textbf{Overview.} We present a proof-of-concept study on improving the sample efficiency of real-world reinforcement learning (RL) by leveraging a simulation-pretrained policy and a digital twin. Specifically, we focus on constrained environments with fixed objects and backgrounds, where we pretrain a task-specific policy within the corresponding digital twin.  
Building on this foundation, we propose a hybrid RL framework based on two key designs: (1) utilizing rollouts from the pretrained policy in both simulation and the real world to bootstrap the critic’s learning, and (2) employing the pretrained policy to propose alternative actions for online exploration and critic bootstrapping, analogous to IBRL~\cite{hu2023imitation}.  
Section~\ref{subsec: sim} details the digital twin reconstruction process and visuomotor policy training. 
In Section~\ref{subsec: simlauncher}, we outline our demonstration collection strategies and real-world RL pipeline, which integrates the pre-trained policy and digital twin. 
We evaluate our approach on a diverse set of manipulation tasks, including \textbf{Pick and Place}, \textbf{Pick and Insert}, and \textbf{Dex Grasp}, as illustrated in Fig.~\ref{fig:task_illustration}.

%%%%%%%%%%% figure tasks %%%%%%%%%%%%
%%%%%%%%%%%%%%%%%%%%%%%%%%%%%%%%%%%%%%

%%%%%%%%%%%%%%%%%%%%%%%%%%%%%%%%%%%%%%
\vspace{-8pt}
\subsection{Simulation Policy Learning and Demo Collection}
\label{subsec: sim}
\textbf{Digital Twin Environments.}
We initially construct digital twin environments for real-world tasks to enhance the sim-to-real transferability of subsequently trained visuomotor policies.
For \textbf{Pick and Place} and \textbf{Pick and Insert}, we employ 3D Gaussian Splatting (3DGS) to reconstruct the geometry and texture of robots and task-related objects while using Isaac Gym~\cite{makoviychuk2021isaac} for physics simulation. 
To render visual observations under different robot and object states, we follow~\cite{qureshi2024splatsim} and transform the objects' Gaussian kernels based on their poses. 
% Due to page limitations, 
We provide further details in the supplementary materials.
For \textbf{Dex Grasp}, we use Mujoco~\cite{todorov2012mujoco} for both physics simulation and visual rendering.

\textbf{Pre-training Visuomotor Policy in Simulation.}  
Our visuomotor policy is pre-trained using a distillation-based approach, following PartManip~\cite{geng2023partmanip}. For each task, we first train a state-based policy with privileged information (proprioceptive data and object states) using RL. This policy generates successful trajectories in simulation, which serve as visual demonstrations. We then apply behavior cloning (BC) to these demonstrations to train a visuomotor policy that takes RGB images and robot proprioceptive data as input.

\textbf{Mitigating the Sim-to-Real Gap.}
For the \textit{physical} sim-to-real gap, we first calibrate physical parameters, camera pose, and robot controller by rolling out the same sequence of actions in both the simulator and the real world and then comparing the trajectories.
For the \textit{visual} sim-to-real gap, we apply the following strategies:
(1) During simulated demonstration collection, we randomize the camera pose within a small range to simulate camera extrinsic calibration errors.
(2) When training the behavior cloning policy, we use random cropping and color jittering as data augmentation.
(3) We mask out the background in both simulated and real-world observations. 
For real-world policy rollout and RL training, we annotate the first frame’s observations to generate initial segmentation masks using SAM2~\cite{ravi2024sam2}. 
Leveraging SAM2’s efficient mask-tracking capability, subsequent masks can be obtained within 0.05s, making it compatible with our 10Hz control frequency.

\subsection{Real-World Online RL with Simulation Bootstrapping}
\label{subsec: simlauncher}
SimLauncher is a simple and effective approach that leverages a simulation-pretrained policy to perform rollouts in both simulation and the real world, enabling the bootstrapping of the critic. Additionally, it utilizes the pre-trained policy to propose actions for online exploration. The corresponding pseudo-code is provided in Appendix~\ref{sec:pseudo_code}.
The key design choices are detailed below:

\textbf{Simulated Demo Collection.} 
Simulation provides a safe and efficient environment for data collection. 
We leverage simulation rollouts for critic bootstrapping, denoted as $\mathcal{D}_{sim}$. Compared to limited real-world demos from policies or humans, simulated demos offer broader coverage of initial conditions and intermediate states.
Our base approach relies on \textit{success-only} demos, but we also explore an extension: collecting rollouts uniformly during state-based policy training with post-rendered image observations, referred to as \textit{hybrid} demos. With better state coverage, hybrid demos improve performance, as discussed in \ref{subsec: analysis}.

\textbf{Real-world Demo Collection.}  
Relying solely on simulated demos can lead to the value underestimation of real-world transitions, negatively impacting online RL performance, as noted by \cite{zhou2024efficient}. To mitigate this, we deploy the BC policy in the real world to collect successful trajectories, denoted as $\mathcal{D}_{real}$.  
To balance the disparity between simulated and real-world demos, we sample equally from $\mathcal{D}_{sim}$ and $\mathcal{D}_{real}$. Additionally, we incorporate the latest successful online trajectories into the real demo buffer to minimize the distribution gap between the replay buffer and the demo buffer.  
Each training batch is composed of 25\% data from $\mathcal{D}_{sim}$, 25\% from $\mathcal{D}_{real}$, and 50\% from the replay buffer $\mathcal{R}$.

\textbf{Action and Bootstrap Proposal.} 
Following IBRL~\cite{hu2023imitation}, we incorporate the behavior cloning policy during online interaction for better exploration. 
Specifically, we select either the BC action $\pi^{bc}$ or the RL action $\pi^{rl}$ by sampling from a Boltzmann distribution over Q-values, i.e., $p_Q(a) \propto \exp(\beta Q(s,a))$ for $a \in \{a^{bc}, a^{rl}\}$, where $\beta$ is the inverse temperature that controls the sharpness of the distribution.
The proposed action is also used for critic bootstrapping. This alters the value target in Eq.~\ref{eq:q-target} to: 
\vspace{-2pt}
\begin{equation}
    y = r + \gamma \underset{a' \in \{a^{rl}, a^{bc}\}}{max}
    Q_{\tilde{\varphi}}(s', a').
\end{equation}

\section{Experiments} 
\label{sec:experiments}

Our experiments aim to evaluate SimLauncher’s effectiveness in improving the sample efficiency of real-world RL for learning a visuomotor policy.
We use a diverse test suite of tasks that challenge online exploration. 
Specifically, we focus on three key questions:
(1) Can SimLauncher, which leverages simulated demonstrations for bootstrapping and a pre-trained policy for exploration, achieve better sample efficiency than the current state-of-the-art hybrid RL methods that rely on human demonstrations?~(Sec.~\ref{sec:baseline})
(2) How crucial are SimLauncher’s key design choices, including simulated demos, a limited number of sim-to-real demos, and the action proposal module?~(Sec.~\ref{sec:ablation})
(3) Why can simulation pre-trained policy benefit real-world RL despite the presence of the sim-to-real gap?~(Sec.~\ref{sec:analysis})

%%%% tables

%%%%%%%%%%%%%%%%%%%%%%%%%%%%%%%%%%%%
\begin{figure*}[t]
\begin{center}
% \vspace{-15pt}
\includegraphics[width=0.90\linewidth, trim=0cm 0.2cm 0cm 0cm, clip]{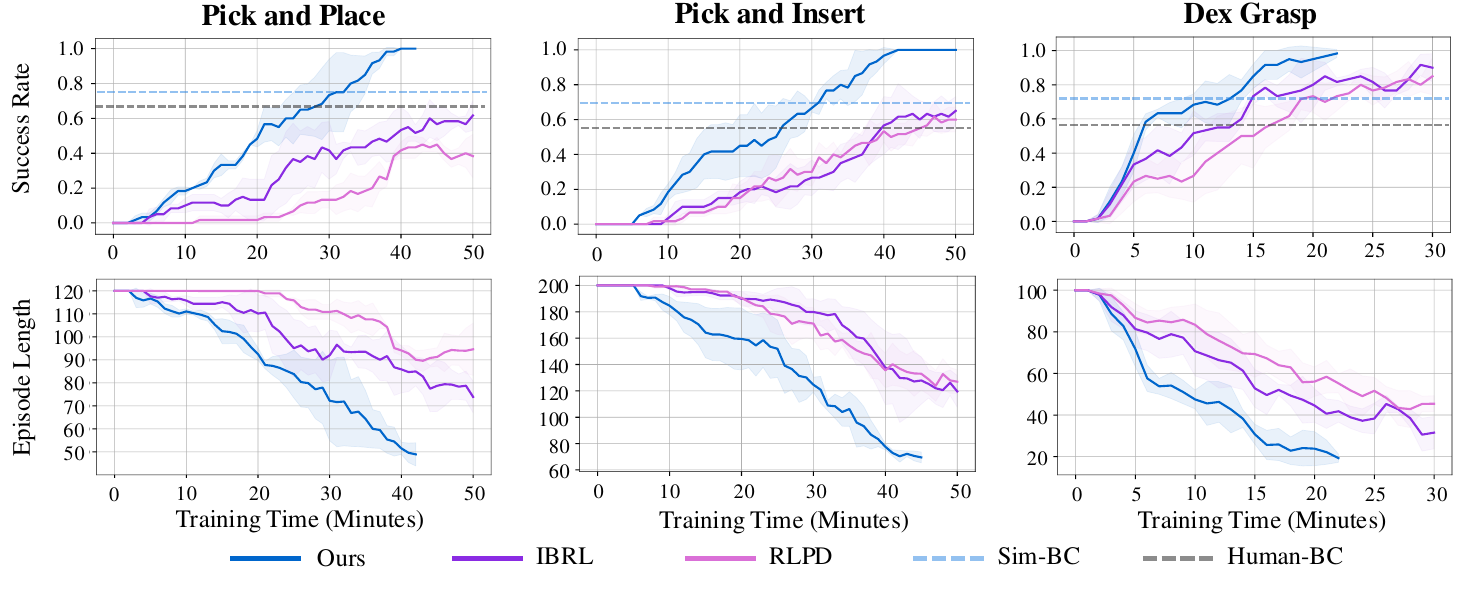}
\end{center}
\vspace{-12pt}
\caption{
\textbf{Comparison with baselines.} 
% We compare our methods with state-of-the-art RL approaches that leverage human-collected demos and behavior cloning methods. 
SimLauncher significantly outperforms state-of-the-art RL approaches that leverage human-collected demos and behavior-cloning methods. We report the mean and standard deviation over 3 seeds.
}
\label{fig:main_results}
\vspace{-15pt}
\end{figure*}
% \input{tabs/main_results_vericle}
%%%%%%%%%%%%%%%%%%%%%%%%%%%%%%%%%%%%%%

\vspace{-5pt}
\subsection{Experimental Setup} 
\label{subsec: exp setup}
\textbf{Tasks and Hardware Setup.} We evaluate online exploration using a diverse suite of tasks. While autonomous reset and reward assignment are key goals for real-world RL, they are beyond this paper's scope. To reduce confounding factors, we manually reset the object and assign a binary reward upon success. 
We set the target control frequency for all tasks to 10Hz.
Figure~\ref{fig:task_illustration} illustrates the task setups, randomization ranges, success criteria, and common failure modes. Details for each task are as follows:

\textbf{Pick and Place:} 
A daily multi-stage robotic manipulation task~\cite{vuong2023open}. The setup features a Franka arm with a Franka Hand parallel gripper. Observations include two third-person RGB images and the gripper state. The action space comprises a 3-DoF delta TCP translation and a binary gripper action. The object is randomly initialized in each episode.

\textbf{Pick and Insert:} 
Another multi-stage task involves contact-rich manipulation, which adds complexity. The hardware setup, observation, and action space match those of Pick and Place. The object is randomly initialized in each episode.

\textbf{Dexterous Hand Grasping:} 
Grasping a columnar can requires leveraging both support and friction forces for stability. We use a Franka arm with a Leap Hand, freezing 5 of its 16 DoFs to prevent excessive finger collisions. Observations include a third-person camera image and hand joint state, while the action space consists of a 3-DoF delta TCP translation and 11-DoF delta hand joint. The object and hand wrist pose are randomly initialized per episode.

% File: table_example.tex
\ifdefined\isMainDocument
% If included in a main document, skip the preamble.
\else
\documentclass{article}

\usepackage{array}
\usepackage{multicol}
\usepackage{multirow}
\usepackage{graphicx}
\usepackage{makecell}
\usepackage{booktabs}
\definecolor{Blue}{HTML}{e5f4fc}
% \definecolor{ColorB}{HTML}{f4e3f4}

\newcommand\mydata[2]{$#1_{\pm#2}$}
\begin{document}

\fi

% Table content
\begin{table}[t]
\centering
\renewcommand{\arraystretch}{1.5} % 调整行高    
\begin{tabular}{c | c | c c c}
  % \hline
  \toprule
  \multirow{2}{*}{Task} & Training & \multicolumn{3}{c}{Success Rate (\%)}  \\
  & Time (min) & Ours & IBRL & RLPD   \\
  % \hline
  \midrule
  Pick and Place & \cellcolor{Blue}$37.5 _{\pm 5.3}$ &\cellcolor{Blue}100.0 & $58.3_{\pm 10.3}$ & $20.0_{\pm 8.2}$   \\
  Pick and Insert & \cellcolor{Blue}$41.8 _{\pm 2.4}$ &\cellcolor{Blue}100.0 & $53.3_{\pm 14.3}$ & $46.7_{\pm 8.5}$   \\
  Dex Grasp & \cellcolor{Blue}$20.0 _{\pm 2.7}$ & \cellcolor{Blue}100.0 & $81.7_{\pm 2.4}$ & $65.0_{\pm 10.8}$   \\
  \bottomrule

\end{tabular}

\caption{\textbf{Evaluation of RL-based methods.} The success rates are reported over 20 trials. For baselines, we evaluate the checkpoints when Ours achieve 100\% success rate. We report the mean and standard deviation over 3 seeds.}
\label{table:main}
\vspace{-17pt}
\end{table}

\ifdefined\isMainDocument
% % If included in a main document, end here.
\else
\end{document}
\fi

\textbf{Baselines.}  
We do not have a strict baseline that perfectly matches our setting.  
Nevertheless, we compare our method with state-of-the-art RL approaches~\cite{hu2023imitation,ball2023efficient} that integrate human-collected offline data with online RL.  
% \begin{itemize}  
% \item \textbf{RLPD} is the first RL algorithm that incorporates an offline prior dataset with online RL. It performs 50/50 sampling from the demonstration buffer and the replay buffer to bootstrap the critic, demonstrating superior results compared to offline-to-online RL methods.  
% \item
\textbf{RLPD} is our core online RL algorithm, as detailed in \ref{subsec: overview}, demonstrating superior results compared to offline-to-online RL methods. 
% \item 
\textbf{IBRL} further incorporates a policy obtained via behavior cloning from the human demos and utilizes the BC policy for action and bootstrap proposal, as detailed in \ref{subsec: simlauncher}. 
IBRL achieves state-of-the-art performance among hybrid RL approaches. Therefore, we do not compare with other hybrid RL algorithms in this work.  
% \end{itemize}  
Following~\cite{luo2024serl}, we provide 20 human demonstrations for IBRL and RLPD, which is a commonly used and user-affordable setting. 

\vspace{-7pt}
\subsection{Comparison with Baselines}
\label{sec:baseline}
To answer Question 1, we compare our method with baselines across three tasks illustrated in Fig.~\ref{fig:task_illustration}. 
% For each task environment, we scan the object and construct the digital twin, except for Dexterous Hand Grasping, where we use Mujoco-rendered RGB images via rasterization as observations. (repeat the method)
While our setup does not perfectly match the baselines, we ensure a fair comparison by setting the number of real demonstrations in SimLauncher to 20, identical to the baselines using human demos. 
% We implement IBRL based on our RLPD implementation
Additionally, we align the usage of offline data in IBRL with our approach for fair comparison, rather than using buffer initialization as in the original paper.
We present learning curves for all tasks in Fig.~\ref{fig:main_results}, where success rate and episode length are computed as running averages over the latest 20 episodes. 
Each experiment is \textbf{repeated 3 times} with different seeds, and we report the mean and variance. 
We also compare the earliest training checkpoint where our method reaches 100\% success with baseline checkpoints at the same timestep, as shown in Tab.~\ref{table:main}.

As illustrated in Fig.~\ref{fig:main_results}, our method consistently outperforms all baselines across all tasks. 
When our method converges to near-perfect performance (100\%), the strongest baseline, IBRL, lags behind by 20–40\% (Tab.~\ref{table:main}). 
These results demonstrate that our approach significantly improves sample efficiency compared to conventional hybrid RL methods that rely on human-collected data.
Notably, on \textbf{Pick and Place} and \textbf{Pick and Insert}, which involve more sequential stages than \textbf{Dex Grasp}, our method exhibits a greater advantage, as shown in Fig.~\ref{fig:main_results} and Tab.~\ref{table:main}. 
This aligns with our intuition that the stage coverage provided by simulation pre-training is particularly beneficial for tasks with more complex exploration challenges.
Surprisingly, both our method and baselines converge within 30 minutes on \textbf{Dex Grasp}, even faster than on \textbf{Pick and Place}. 
This may be due to a higher FPS of the actor and learner nodes during training, reaching 10 Hz and 12 Hz, respectively. 
However, our method requires 16000 learner steps to converge on \textbf{Dex Grasp}, which exceeds the 10000 steps needed on \textbf{Pick and Place}.

%%%%%%%%%%% figure ablations %%%%%%%%%%%%
\begin{figure}[t]
\begin{center}
\includegraphics[width=0.9\linewidth, trim=0.1cm 0cm 0cm 0.2cm, clip]{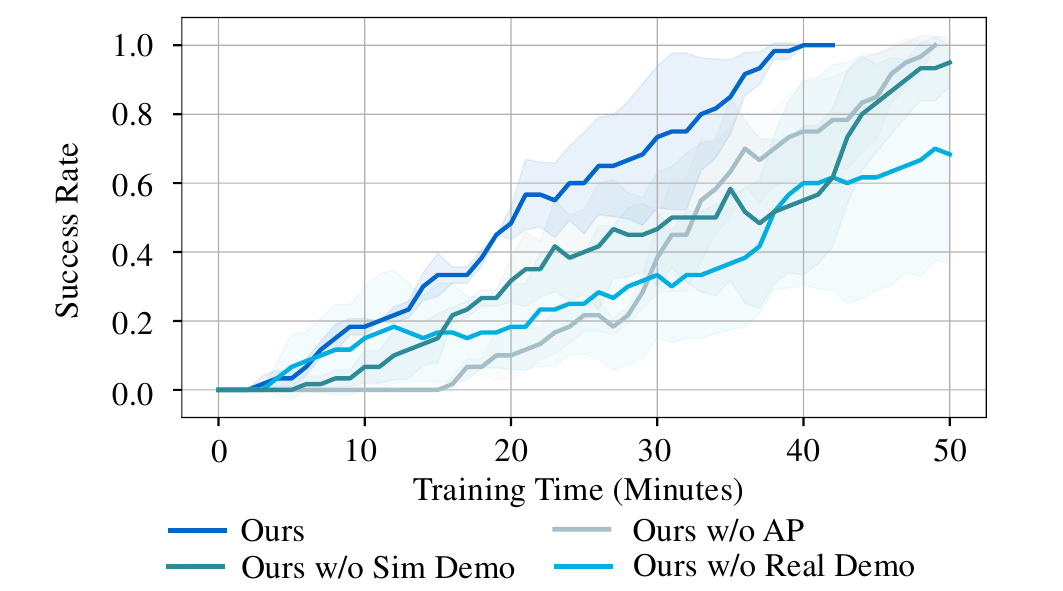}
\end{center}
\vspace{-7pt}
\caption{Ablation study on our key design choices, 3 seeds. }
\label{fig:ablation}
\vspace{-20pt}
\end{figure}

\vspace{-5pt}
\subsection{Ablation Studies}
\label{sec:ablation}

To answer Question 2, we compare our method with three ablated versions: \textbf{Ours w/o Sim Demo}, \textbf{Ours w/o Real Demo}, and \textbf{Ours w/o AP}, which respectively remove simulated demos, sim-to-real rollout demos, and action proposals by the simulation-pretrained policy, on \textbf{Pick and Place}. 
The evaluation follows the same procedure as in Sec.~\ref{sec:baseline}.

As shown in Fig.~\ref{fig:ablation}, all three ablated versions underperform the full method, underscoring the importance of each design choice. 
\textbf{Ours w/o Real Demo} suffers the most significant drop, likely due to critic overfitting, emphasizing the necessity of real-world demos for regularization. 
Its critic assigns lower values to real-world interactions, possibly because bootstrapping from both successful simulated demos and low-success-rate replay exacerbates overfitting.
\textbf{Ours w/o AP} struggles early on, as other methods leverage a pre-trained policy with a 73.3\% success rate. 
However, after the ``cold start,'' it rapidly improves and converges faster than \textbf{Ours w/o Real Demo} and \textbf{Ours w/o Sim Demo}, likely due to the combined benefits of simulated and real-world demos enhancing stage coverage and stabilizing training.

\vspace{-7pt}
\subsection{Analysis}
\vspace{-3pt}
\label{sec:analysis}
\label{subsec: analysis}
To answer Question 3, we study the robustness of our proposed action proposal and simulated demonstration bootstrapping approach to the sim-to-real gap. Below all the experiments are conducted on \textbf{Pick and Place}.

\textbf{Takeaway 1: Scaling simulation pre-training improves sim-to-real policy transfer.}
We compare a BC policy trained on different numbers of simulated demos with one trained on human-collected demonstrations. As shown in Tab.  ~\ref{table:bc scaling}, the performance of the BC policy on simulated data consistently improves as the training data increases. Although Sim-BC underperforms Human-BC when trained on the same amount of data (e.g., 20 trajectories), likely due to the sim-to-real gap, simulation allows for scalable data collection. 
% By significantly increasing the amount of training data (e.g., to 1000 trajectories), Sim-BC can achieve a high success rate in the real world despite the sim-to-real gap.
With extensive training (e.g., 1000 trajectories), Sim-BC achieves high real-world success despite the sim-to-real gap.

% File: table_example.tex
\ifdefined\isMainDocument
% If included in a main document, skip the preamble.
\else
\documentclass{article}

\usepackage{array}
\usepackage{multicol}
\usepackage{multirow}
\usepackage{graphicx}
\usepackage{makecell}
\usepackage{booktabs}

\newcommand\mydata[2]{$#1_{\pm#2}$}
\begin{document}

\fi

% Table content
\begin{table}[ht]
\vspace{-8pt}
\centering
\renewcommand{\arraystretch}{1.5} % 调整行高    
\begin{tabular}{c|c| c c c c c}
  \toprule
  & Human & \multicolumn{5}{c}{Simulation} \\
  \midrule
  \# of Demos & 20 & 10 & 20 & 50 & 100 & 1000 \\
  \midrule
  Success Rate (\%)& 65 & 25 & 45 & 55 & 70 & 75 \\
  \bottomrule
\end{tabular}
\vspace{-3pt}
\caption{\small Scaling simulation data for behavior cloning.}
\label{table:bc scaling}
\vspace{-10pt}
\end{table}

\ifdefined\isMainDocument
% % If included in a main document, end here.
\else
\end{document}
\fi

%%%%%%%%%%% figure sim demo rlpd %%%%%%%%%%%%
\begin{figure}[t]
\begin{center}
\includegraphics[width=0.9\linewidth, trim=0cm 0.6cm 0cm 0.1cm, clip]{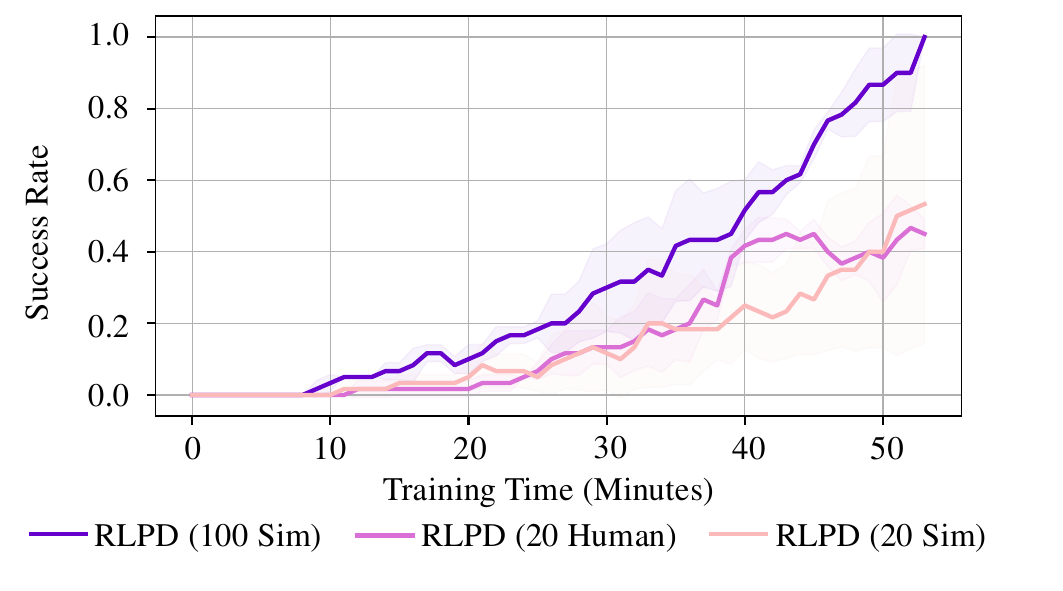}
\end{center}
\vspace{-10pt}
\caption{\footnotesize Simulated demos alone enable effective bootstrapping, 3 seeds.}
\label{fig:Sim demo RLPD}
\vspace{-20pt}
\end{figure}

\textbf{Takeaway 2: Simulated demonstrations alone enable effective bootstrapping.}
A key concern when using simulated demos for bootstrapping is that the critic might overfit to simulation-specific features, allowing it to distinguish between simulated demos and the real-world replay buffer. 
To examine this, we compare the RLPD method using different sources of demos: 100 simulated demos, 20 human-collected real demos, and 20 simulated demos. 
As shown in Fig.~\ref{fig:Sim demo RLPD}, simulated demos alone can effectively bootstrap real-world RL. While RLPD with 20 simulated demos performs slightly worse than using 20 human demos, scaling up the number of simulated demos to 100 significantly improves efficiency, surpassing RLPD with 20 human demos.

%%%%%%%%%%% figure hybrid demo %%%%%%%%%%%%
\begin{figure}[t]
\begin{center}
\includegraphics[width=0.9\linewidth, trim=0cm 0.6cm 0cm 0.1cm, clip]{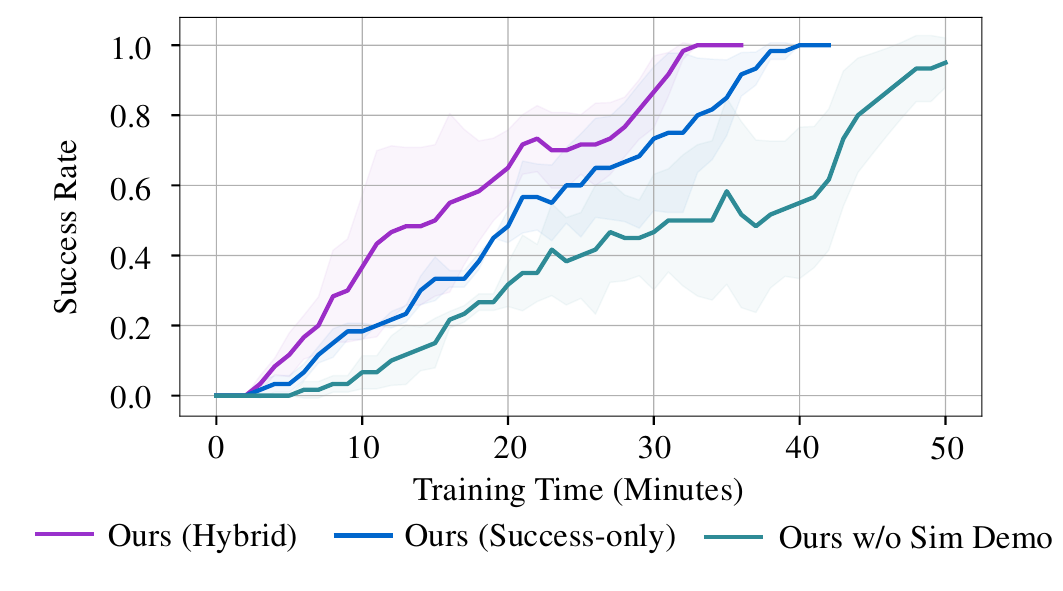}
\end{center}
\vspace{-10pt}
\caption{\small Enlarging state-coverage of simulated demonstrations can further improve sample-efficiency of \textbf{Ours}, 3 seeds. }
\label{fig:Hybrid demo}
\vspace{-10pt}
\end{figure}

%%%%%%%%%%% new
\begin{figure}[h]
\begin{center}
% \vspace{-10pt}
\includegraphics[width=\linewidth, trim=0cm 0cm 0cm 0cm, clip]{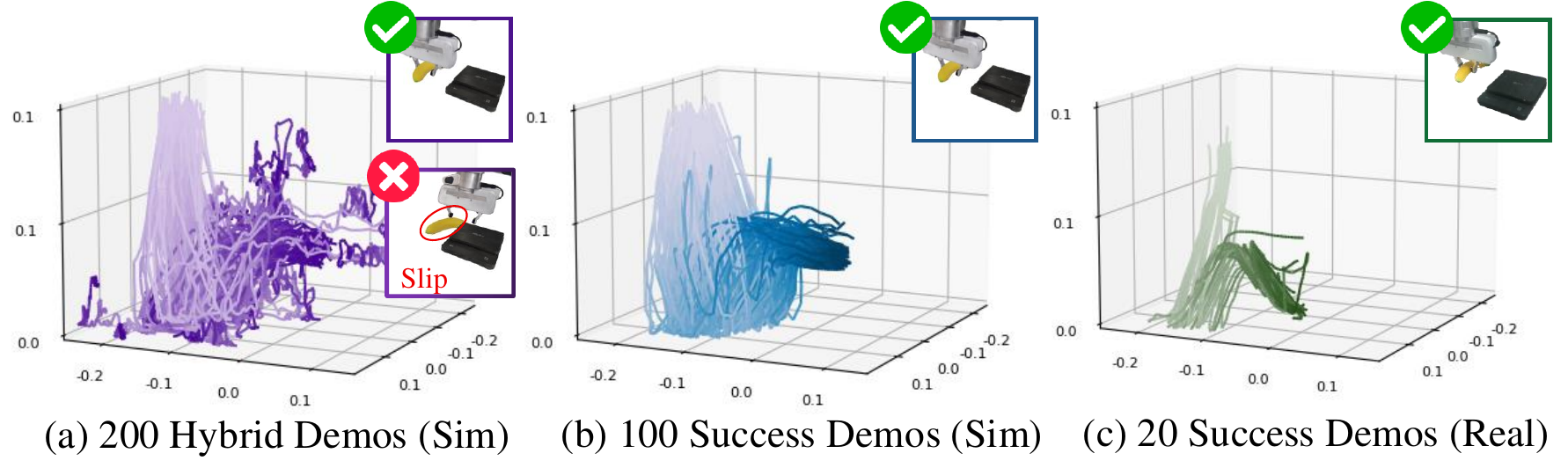}
\end{center}
\vspace{-10pt}
    \caption{\small Visualization of state coverage in hybrid and success-only simulated demonstrations on the Pick and Place task.}
    \label{fig:state_coverage}
\vspace{-20pt}
\end{figure}

\textbf{Takeaway 3: Expanding state coverage in simulated demonstrations enhances bootstrapping.}
A key advantage of simulation is its scalability in data generation. 
This raises the question of whether increasing state coverage in simulated demonstrations can improve sample efficiency. 
% To investigate this, we construct a ``hybrid demo buffer" by sampling 50 failure trajectories and 150 success trajectories from the replay buffer used for pre-training the simulation policy. 
To investigate this, we construct a ``hybrid demo" by collecting rollouts uniformly throughout the state-based policy training process, and post-render the image observations.
As illustrated in Fig.~\ref{fig:state_coverage}, this approach provides better state coverage. 
Furthermore, as shown in Fig.~\ref{fig:Hybrid demo}, using the hybrid demo buffer leads to slightly improved sample efficiency, suggesting that expanding state coverage in simulated demonstrations can further enhance bootstrapping.

\vspace{-3pt}
\section{Conclusion} 
\vspace{-3pt}
\label{sec:conclusion}
We present SimLauncher, a vision-based real-world reinforcement learning approach that integrates digital twins to bridge simulation pretraining and real-world policy optimization. 
Simlauncher leverages simulated and real-world rollouts from the simulation pre-trained policy for critic bootstrapping and combines action proposals from the pre-trained policy for better exploration. 
Our method achieves superior sample efficiency compared to conventional hybrid RL approaches with an affordable scale of human demonstration across multi-stage tasks, precision manipulation challenges, and high-DoF dexterous manipulation.
% through extensive ablations and analysis, we prove that xxxxxx despite the sim-to-real gap.

\textbf{Limitations and Future Works.}
SimLauncher's adaptability is fundamentally constrained by the simulation environment, which struggles to accurately replicate highly dynamic scenarios, high-precision tasks, or interactions with deformable objects. Our current implementation relies on real-time segmentation for 
% both simulated and 
real-world observations. 
In future work, we may leverage large-scale training or domain randomization to reduce this dependency. Additionally, this proof of concept does not include a fully autonomous system for self-resetting or reward control; future works might incorporate advances in autonomous reinforcement learning, such as training a reward classifier~\cite{luo2024serl} and a reset policy~\cite{hu2023reboot}.

% real-to-sim engineering effort


% \section*{ACKNOWLEDGMENT}

% The preferred spelling of the word ÒacknowledgmentÓ in America is without an ÒeÓ after the ÒgÓ. Avoid the stilted expression, ÒOne of us (R. B. G.) thanks . . .Ó  Instead, try ÒR. B. G. thanksÓ. Put sponsor acknowledgments in the unnumbered footnote on the first page.
\newpage
\appendix
\subsection{PSEUDO CODE of SimLauncher}
\label{sec:pseudo_code}
\begin{algorithm}[]
  \caption{SimLauncher}
  \begin{algorithmic}
    \STATE \textbf{Hyperparameters}: Entropy Temperature $\alpha$, Action Proposal Inverse Temperature $\beta$, Gradient Steps $G$
    \STATE Randomly initialize Critic $\varphi_i$ (set targets $\varphi' = \varphi$) and Actor $\theta$ parameters.
    \STATE Initialize empty replay buffer $\mathcal{R}$
    \STATE \textcolor{LightBlue}{Initialize real demo buffer $\mathcal{D}_{real}$ with real-world demos and sim demo buffer $\mathcal{D}_{sim}$ with simulated demos}
    \WHILE{True}
      \STATE Receive initial observation state $s_0$
      \FOR{t = 0, T}
        \STATE Compute action $a_t^{rl} \sim \pi_\theta(\cdot|s_t) $ and $a_t^{bc}\sim\pi_{bc}(\cdot|s_t)$
        \STATE Take action $a_t \sim \text{softmax}_{ a \in \{a^{bc}, a^{rl}\}}(\beta Q(s,a))$
        \STATE Store transition $(s_t, a_t,
                r_t, s_{t+1})$ in $\mathcal{R}$
        \FOR{{$g = 1, G$}}
            \STATE Sample minibatch $b_R$ of {$\frac{N}{2}$} 
            from $\mathcal{R}$
            \STATE \textcolor{LightBlue}{Sample minibatch $b_{D_{real}}$ 
            of $\frac{N}{4}$ 
            from $\mathcal{D}_{real}$}
            \STATE \textcolor{LightBlue}{Sample minibatch $b_{D_{sim}}$ 
            of $\frac{N}{4}$ 
            from $\mathcal{D}_{sim}$}
            \STATE {Concatenate $b_R$, $b_{D_{real}}$ and $b_{D_{sim}}$ to form batch $b$ of size $N$}
            set
            \STATE {For element $j$ in  $b$}
            \begin{equation}
                y^{j} = r^{j} + \gamma \underset{a' \in \{a^{rl}, a^{bc}\}}{max}
                Q_{\tilde{\varphi}}(s', a').
            \end{equation} 
            % \vspace{-2mm}
            \STATE Update $\varphi$ minimizing loss:
                % \vspace{-2mm}
                \begin{equation}
                   \mathcal{L}_Q(\varphi) = \frac{1}{N}\sum_{j}\left( Q_{\varphi}(s, a) - y\right)
                \end{equation}
                % \vspace{-4mm}
            \STATE Update target networks $\varphi_i' \leftarrow \rho\varphi_i' + (1-\rho)\varphi_i$
        \ENDFOR
        \STATE Update $\theta$ maximizing objective:
        % \vspace{-2mm}
        \begin{equation}
        % \small
            \frac{1}{N}\sum_{j} Q_{\varphi}(s^{j}, \pi_\theta(s^j)) - \alpha \,\text{log}\,\pi_\theta(\pi_\theta(s^j)|s^j),
        \end{equation}
        % \vspace{-3mm}
        \ENDFOR
        \IF{\textcolor{LightBlue}{trajectory succeed:}}
            \STATE \textcolor{LightBlue}{Append trajectory into $\mathcal{D}_{real}$}
        \ENDIF
    \ENDWHILE
  \end{algorithmic}
\end{algorithm}

\subsection{Implementation Details}
\paragraph{Shared setting}
We set the following shared settings for the three tasks. Following HIL-SERL, we use DrQ to control the gripper action. Successful trajectories from online rollouts are appended to the real demo buffer. We use a discount factor of 0.97. The inverse temperature will gradually go from an initial number to infinity, the latter is equivalent to argmax.
\paragraph{Pick and Place}
The randomization range for the object is 10cm in x and y. We set the initial inverse temperature to 50, the size of the critic ensemble to 10, and the sub-sample number to 2.
\paragraph{Pick and Insert}
The randomization range for the object is 8cm in x and y. We set the initial inverse temperature to 50, the size of the critic ensemble to 1/0, and the sub-sample number to 2.
\paragraph{Dex Grasp}
The randomization range for the object is 10cm in the xy plane and 4cm in x, y, and z for the hand. We set the initial inverse temperature to 10, the size of the critic ensemble to 10, and the sub-sample number to 10.


\bibliographystyle{IEEEtran}
\bibliography{references}

% \section{Appendix}

\end{document}